\documentclass[journal]{IEEEtran}
\usepackage{graphicx}
\usepackage{caption}
\usepackage{float}
\usepackage{subcaption}

\usepackage[linesnumbered,ruled]{algorithm2e}
\usepackage{color}
\usepackage[table]{xcolor}
\usepackage{url}
\usepackage[colorinlistoftodos]{todonotes}

\def\PPAMI{{\textsf{M2P2}}} 
\newcommand{\norm}[1]{\left\lVert#1\right\rVert}

\newcommand{\edit}[1]{{#1}}
\newcommand{\editrow}{}

\newcommand{\nop}[1]{{}}

\usepackage{amsmath}

\ifCLASSINFOpdf
\else
\fi

\begin{document}

\title{M2P2: Multimodal Persuasion Prediction using Adaptive Fusion}

\author{
Chongyang Bai,
Haipeng Chen,
Srijan Kumar,
Jure Leskovec,
and V.S. Subrahmanian
\thanks{C. Bai is with the Department of Computer Science, Dartmouth College, Hanover, NH 03755, USA. H. Chen is with the Department of Computer Science, Harvard University, Boston, MA 02138, USA. S. Kumar is with the College of Computing at Georgia Institute of Technology, Atlanta, GA 30332, USA. J. Leskovec is with the Department of Computer Science at Stanford University, Stanford, CA 94305, USA. V.S. Subrahmanian is with the Department of Computer Science and the Roberta Buffett Institute of Global Affairs at Northwestern University, Evanston, IL 60208, USA.}
\thanks{Corresponding author: Professor V.S. Subrahmanian.}
\thanks{Code and datasets can be found at https://snap.stanford.edu/persuasion.}
}

\markboth{IEEE Transactions on Multimedia}%
{Shell \MakeLowercase{\textit{et al.}}: Bare Demo of IEEEtran.cls for IEEE Journals}

\renewcommand\thefigure{\arabic{figure}}
\setcounter{figure}{0}

\makeatletter
\let\@oldmaketitle\@maketitle
\renewcommand{\@maketitle}{\@oldmaketitle
    \begin{center}
    \setcounter{figure}{0}
    \vspace{2mm}
    \includegraphics[width=\textwidth]{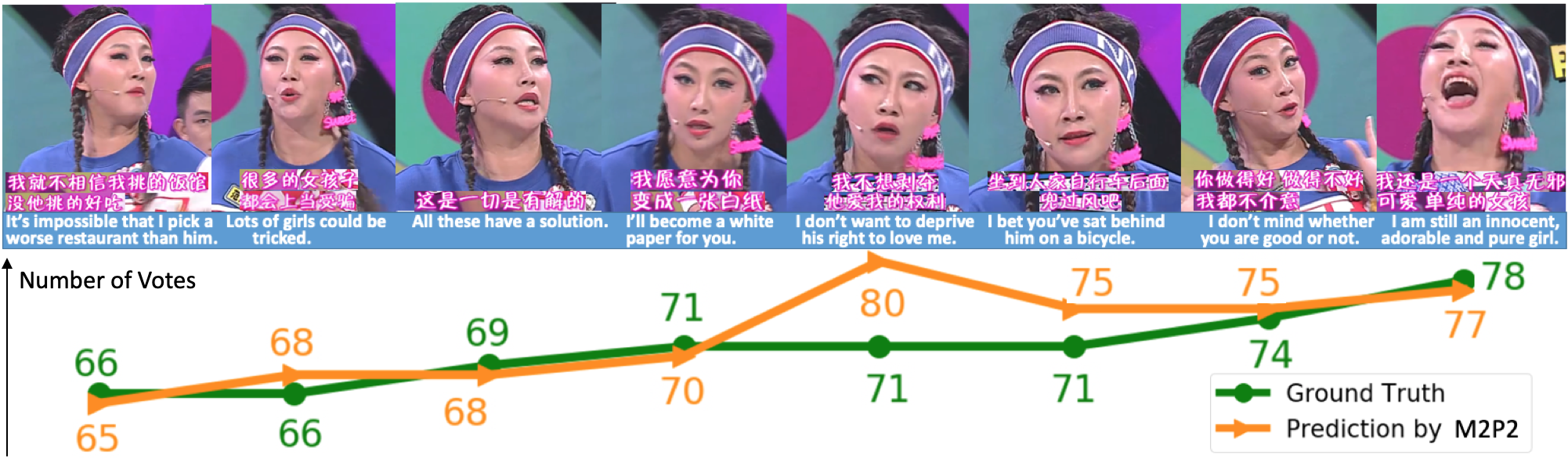}
    \captionof{figure}{Real-time prediction of debate persuasiveness (number of votes) using our proposed model \PPAMI. The debate is from a Chinese debate TV show, Qipashuo. \PPAMI\ closely predicts the ground truth number of votes.}\label{fig:real_time_prediction}
    \end{center}
    }
\makeatother

\maketitle


\begin{abstract}
Identifying persuasive speakers in an adversarial environment is a critical task. In a national election, politicians would like to have persuasive speakers campaign on their behalf. When a company faces adverse publicity, they would like to engage persuasive advocates for their position in the presence of adversaries who are critical of them. Debates represent a common platform for these forms of adversarial persuasion. This paper solves two problems: the Debate Outcome Prediction (DOP) problem  predicts who wins a debate while the Intensity of Persuasion Prediction (IPP) problem predicts the change in the number of votes before and after a speaker speaks. Though DOP has been previously studied, we are the first to study IPP.  Past studies on DOP fail to leverage two important aspects of multimodal data: 1) multiple modalities are often semantically aligned, and 2) different modalities may provide diverse information for prediction. Our \PPAMI\ (Multimodal Persuasion Prediction) framework is the first to use multimodal (acoustic, visual, language) data to solve the IPP problem. 
To leverage the alignment of different modalities while maintaining the diversity of the cues they provide, \PPAMI\ devises a novel adaptive fusion learning framework which fuses embeddings obtained from two modules -- an \emph{alignment} module that extracts shared information between modalities and a \emph{heterogeneity} module that learns the weights of different modalities with guidance from three separately trained unimodal reference models. We test \PPAMI\ on the popular IQ2US dataset designed for DOP. We also introduce a new dataset called QPS (from Qipashuo, a popular Chinese debate TV show ) for IPP. \PPAMI\ significantly outperforms 4 recent baselines on both datasets. 



\end{abstract}


%
\IEEEpeerreviewmaketitle

\section{Introduction}\label{sec:intro}

Controversial topics (e.g. foreign policy, immigration, national debt, privacy issues) engender much debate amongst academics, businesses, and politicians. Speakers who are persuasive often win such debates. Given videos of discussions between two participants, the goal of this paper is to provide a fully automated system to solve two persuasion related problems. The Debate Outcome Prediction problem (DOP) tries to determine which of two teams ``wins'' a debate. \edit{Suppose the two teams are $A$ and $B$ and suppose
 $bef_A,bef_B$ denote the number of voters for $A$ and $B$'s positions respectively before the debate and $aft_A,aft_B$ denote the same after the debate. Hence, $bef_A+bef_B < n \text{ and } aft_A+aft_B < n$ where $n$ is the total number of voters in the audience. In the DOP problem, we say that team $A$ (resp. team $B$) wins the debate if $aft_A - bef_A > aft_B - bef_B$ (resp. $<$).} We say a speaker is a winner if s/he belongs to the winning team.
 The Intensity of Persuasion Problem (IPP) tries to predict the increase (or decrease) in the number of votes of each speaker (as opposed to a team). We use the same notation as before but assuming we have two speakers $S_1,S_2$. The intensity of speaker $X$'s persuasiveness is $\frac{aft_X-bef_X}{n}$ for $X\in\{S_1,S_2 \}$.
 It is clear that both these problems are important. In a business meeting, it might be important to win (DOP), but in other situations, peeling away support for an opponent might be important (IPP). \emph{The more support a speaker can peel away from the opponent, the more persuasive s/he is.}

\begin{figure}[t]
	\centering
		\includegraphics[width=\columnwidth]{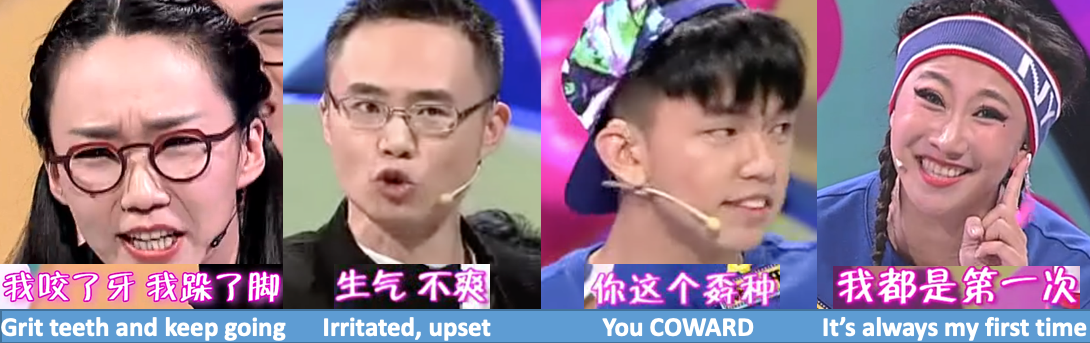}
\caption{\textit{In multimodal content, the modalities are semantically aligned.} This example shows a case where the visual modality (facial expressions) and the language modality (the content of the speech) are closely aligned.}
	\label{fig:alignment}
\end{figure}

\begin{figure}
\centering
\begin{subfigure}[b]{\columnwidth}
\centering
   \includegraphics[width=\columnwidth]{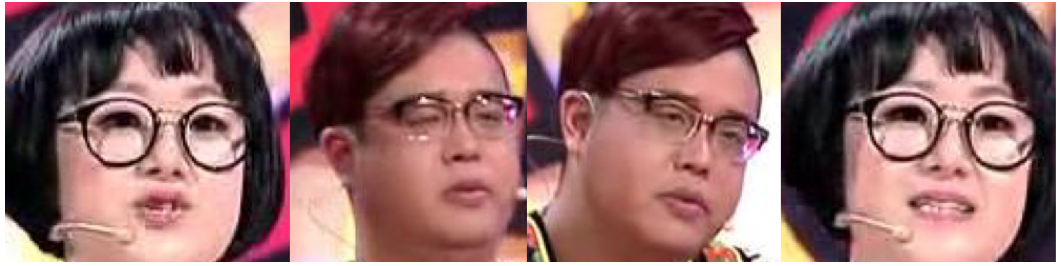}
   \caption{\textit{There are cases where the visual modality is noisy, while the language modality is clean.} In 4 consecutive frames when the woman is speaking, the face of a man appears (see frames 2 and 3) and the man's face is incorrectly assumed to be the woman's. The language modality, however, is correct.}
   \label{fig:noisy_video} 
\end{subfigure}

\hfill\hfill

\begin{subfigure}[b]{\columnwidth}
\centering
   \includegraphics[width=\columnwidth]{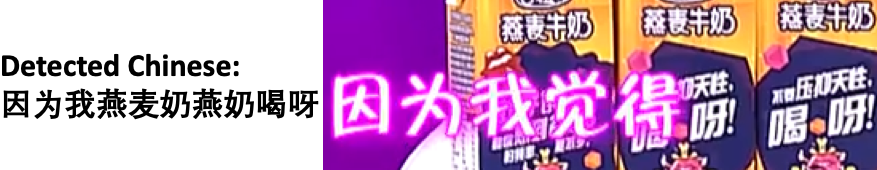}
   \caption{\textit{There are cases where the language modality can be noisy, while the visual modality is clean.} In the video frame (the right side of the figure), the subtitles extracted by the OCR system (the left side) are incorrect due to the milk ads shown. \edit{Moreover, the font and color of subtitles vary from video to video, so it is difficult to automatically separate these subtitles from other texts. }}

   \label{fig:noisy_language}
\end{subfigure}
\caption{\textit{Individual modalities can be noisy.} Here we show examples where the visual or the language modality are wrong. \PPAMI\ learns to down-weight the noisy modalities.}\label{fig:noisy-modality}
\end{figure}

Solving DOP and IPP using video data alone can pose many challenges. In this paper, we test our \PPAMI\ algorithm against two datasets, the IQ2US dataset\footnote{https://www.intelligencesquaredus.org} from a popular US debate TV show and the QPS dataset from the popular Chinese TV show Qipashuo\footnote{\label{qps-url}https://www.imdb.com/title/tt4397792/}.
Real-world videos such as these come with three broad properties:
(i) as we can see in Figure~\ref{fig:noisy_language}, the detected language can be very noisy --- this must be accounted for, (ii) as we can see from Figure~\ref{fig:noisy_video}, there can be considerable noise in the video modality as well --- for instance, a man's face might be shown in the video while a woman is speaking and these kinds of audio-video mismatches must be addressed,  (iii)  but in some cases --- as shown in Figure~\ref{fig:alignment}, the modalities might be nicely aligned where the audio, language, and video modalities are all correct and the speaker's speech and visual signals are aligned. The problem of identifying these types of mismatches poses a major challenge in building a single model to predict both DOP and IPP.

Though we are not the first to take on the DOP problem, we are the first to solve IPP. DOP has been addressed by~\cite{brilman2015multimodal,nojavanasghari2016deep,santos2018multimodal} who use multimodal sequence data to predict who will win a debate.
However, these efforts do not address all the three  challenges described above. 
To the best of our knowledge, there is no existing dataset that addresses the IPP problem and there are no algorithms to solve the IPP problem. 
\emph{In this paper, we develop a novel algorithm called \PPAMI\ and show that \PPAMI\ improves upon past solutions to DOP by 2\%--3.7\% accuracy (statistically significant with a $p$-value below 0.05) and beats adaptations of past work on DOP to the IPP case by over 25\% MSE (statistically significant with $p < 0.01$).}
Figure~\ref{fig:real_time_prediction} shows a sample of how our \PPAMI\ framework predicts speaker persuasiveness at interim points during a debate from the QPS dataset --- the reader can readily see that the \PPAMI\ prediction of number of votes (orange line) closely matches the ground truth (green line).


When all three modalities (audio, video, language) agree, then that ``common'' information must be correctly captured by a predictive model. In this case, we say that the modalities are \emph{aligned}. However, there can be cases where some modalities suggest one thing, while the other(s) suggest something different. In this case, we say the modalities are \emph{heterogeneous}. Our solution, \PPAMI, captures both aspects and also learns how to weight the two aspects in order to maximize prediction accuracy.
 \PPAMI\ first leverages the Transformer encoder structure~\cite{vaswani2017attention} to project the three modalities into three latent spaces. To combine the information from the latent spaces, the model then devises two major modules: \emph{alignment} and \emph{heterogeneity}. 

The \emph{alignment} module learns to highlight the shared, aligned information across modalities. It enforces an alignment loss in the loss function as a regularization term during training.
This ensures that there is relatively little discrepancy between the latent embeddings of different modalities when they are aligned. 

The \emph{heterogeneity} module first learns the weights of modality-specific information and applies weighted fusion to harden the model against noisy modalities (cf. Figure~\ref{fig:noisy-modality}).  \PPAMI\ uses a novel interactive training procedure to learn the  weights from three separately trained reference models, each corresponding to a single modality. Intuitively, a modality with smaller unimodal loss should be assigned a higher weight in the multimodal model.
Finally, the outputs of both modules are combined with the debate meta-data for persuasion prediction.

We evaluate \PPAMI\ on the IQ2US and QPS datasets. IQ2US was first used by~\cite{brilman2015multimodal} to evaluate the DOP problem. The IQ2US dataset only has the final debate outcomes, without any labels about how persuasive each speaker is during the debate. Hence, IQ2US cannot be used to evaluate IPP. To this end, we created a new dataset \emph{QPS}, based on an extremely popular Chinese entertainment debate TV show called Qipashuo\footnotemark[2]. In QPS, the audience provides real-time votes before and after each speaker in order to gauge how persuasive the speaker is. 
QPS therefore provides a direct measure of each speaker's persuasiveness for training and evaluation. 
We use the IQ2US dataset for the DOP problem and the QPS dataset for IPP problem.
\PPAMI\ outperforms baselines based on four recent papers \cite{brilman2015multimodal,nojavanasghari2016deep,santos2018multimodal,verma2019deepcu} which were originally designed to predict debate outcomes (or other related problem scenarios). 
We also conduct ablation studies and visualize  our results to show the effectiveness of different novel components in \PPAMI. 

To summarize, we make the following contributions:
\begin{itemize}
    \item To the best of our knowledge, \PPAMI\ is the first to solve the IPP problem. 
    \item We design a novel adaptive fusion learning framework to  solve the IPP and DOP problems. 
    \item We curate a new dataset QPS from the well-known Chinese debate TV show Qipashuo. QPS will be a strong benchmark for future work on persuasion prediction as well as multimodal learning.
    \item \PPAMI\ outperforms reasonable baselines adapting recent papers \cite{brilman2015multimodal,nojavanasghari2016deep,santos2018multimodal,verma2019deepcu} in IPP and DOP problems --- and these improvements are statistically significant. 
    \end{itemize}

\section{Related Work}

\noindent\textbf{Unimodal persuasion prediction.} There has been some work on using a single modality for predicting persuasion.   \cite{zhang2016conversational,potash2017towards,wang2017winning,habernal2016argument} explored the linguistic modality by studying style, context, semantic features and argument-level dynamics in English transcripts to solve DOP. For the visual modality, Joo et al. \cite{joo2014visual} defined nine visual intents related to persuasion (e.g. dominance, trustworthiness) and trained SVMs to predict them and persuasion using hand-crafted features. Huang et al. \cite{huang2016inferring} improved these results by fine-tuning pre-trained CNNs. In the case of audio, MFCC features and LSTM were used by \cite{santos2016domain} to solve DOP. 

\noindent\textbf{Multimodal persuasion prediction.} Brilman et al.~\cite{brilman2015multimodal} solved DOP by extracting facial emotions, voice pitch and word category related features and then training separate SVMs for each modality. The overall prediction for DOP was obtained through a majority vote by the three models.
Nojavanasghari et al.~\cite{nojavanasghari2016deep} solved DOP by first building a Multi-Layer Perceptron (MLP) for each modality, then concatenating the predicted probabilities,and sending them as input to yet another MLP. Because both methods use simple aggregate feature values (e.g. mean, median), they ignore the dynamics of features over time. As a result, these two approaches do not work well with short video clips, and do not leverage temporal dynamics. To address this problem, Santos et al.~\cite{santos2018multimodal} used an LSTM to take each time-step into account, but their feature-level multimodal fusion considers all modalities to be equally important --- thus ignoring the noise, heterogeneity, and alignment properties.
 
\PPAMI\ is the first to address the Intensity of Persuasion Prediction problem (IPP). Moreover, \PPAMI\ captures temporal dynamics via a multi-headed attention mechanism that: (i) learns the importance of different modalities at different times in long video sequences, and (ii) thus learns better representations of multiple modalities. Moreover, \PPAMI\ is the first to capture both alignment and heterogeneity --- hence addressing noise. With these innovations, \PPAMI\ performs well in both IPP and DOP.

\noindent\textbf{General Multimodal Learning.}  A body of multimodal learning methods defines constraints between modalities in a latent space to capture their inter-relationships. Andrew et al. ~\cite{andrew2013deep}  extended Canonical Correlation Analysis by deep neural networks to maximize inter-modal correlations. Song et al. ~\cite{9127152} further proposed to maximize the correlation of the residual matrices of multimodal features. Such correlation constraints have since been used in human action recognition \cite{8489981}, emotion recognition \cite{aguilar-etal-2019-multimodal} and video captioning \cite{7984828}. In addition to capturing the shared relationship, \cite{panagakis2015robust,7952687,verma2019deepcu} tried to extract the individual component of each modality through low-rank estimation. \cite{jo2019cross,7239600} trained cross-modal encoders to reconstruct a modality from another modality. While these efforts provide important insights for creating multimodal embeddings, they do not show how to combine the learned embeddings for accurate prediction.

A second body of work explores architectures for fusing embeddings from modalities. Zadeh et al. \cite{zadeh-etal-2017-tensor} introduced bimodal and trimodal tensors via cross products to express inter-modal features. Mai et al. \cite{8752006} further proposed to combine local and global interaction learning for time-dependent multimodal fusion. As cross products significantly increase the dimensionality of the feature space, \cite{hadamard17,ben2017mutan,zhang2020beyond} introduced bilinear pooling techniques to learn compact representations. Although these methods explicitly model inter-modal relationships, they introduce additional features that require larger networks to be learned for subsequent prediction tasks. In contrast, attention-based fusion \cite{long2018multimodal,9206083} learns the weighted sum of multimodal embeddings taking the prediction task into account. However, they require huge amounts of data to learn the optimal attention weights. In order to capture long-term dependencies, \PPAMI\ uses the Transformer encoder \cite{vaswani2017attention,tsai2019MULT} to learn latent embeddings for modalities. On one hand, inspired by the first class of work, \PPAMI\ uses a shared projector and enforces high correlation among the encoded embeddings. On the other hand, \PPAMI\ computes a weighted concatenation of latent unimodal embeddings, where the weights are guided by the persuasiveness loss of each embedding through interactive training. These two innovations lead to a compact embedding that can be learned with a small dataset.

\section{The \PPAMI\ Framework}
\begin{figure*}[t]
	\centering
		\includegraphics[width=\textwidth,scale=1]{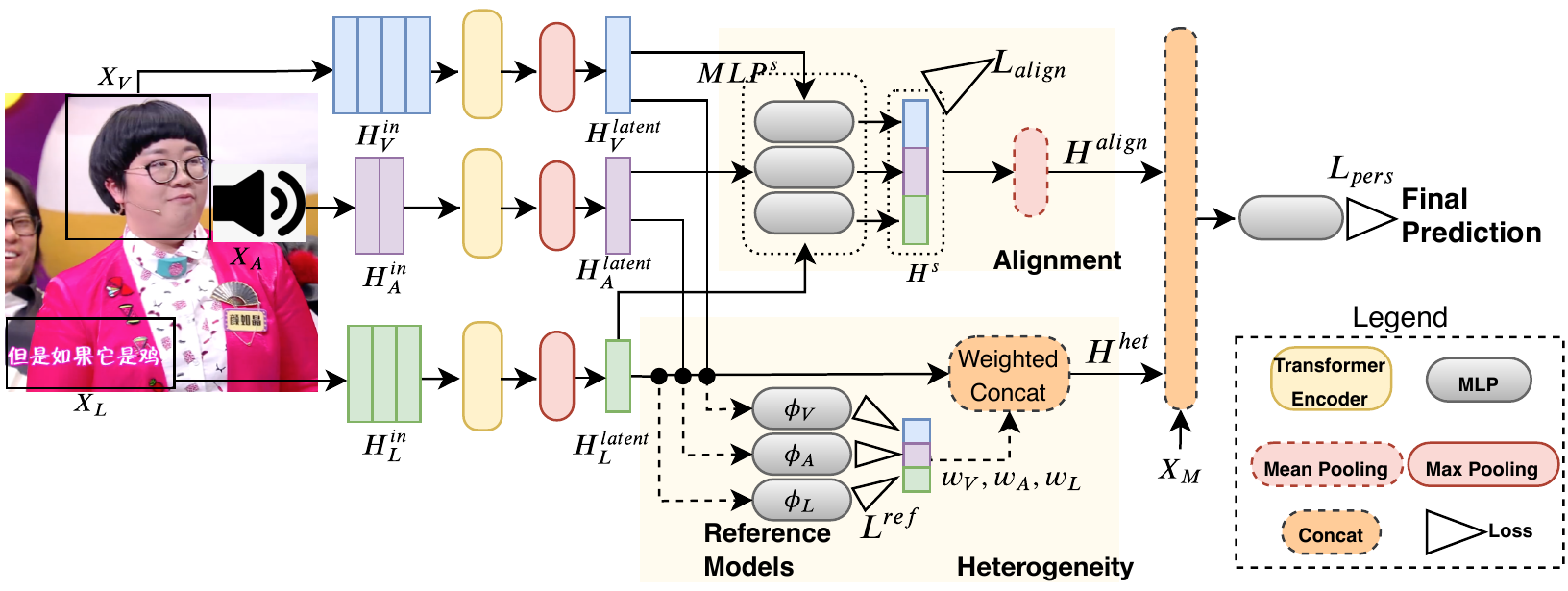}
\caption{\PPAMI\ architecture. First, audio, face and language sequences are extracted from a video clip and fed to three separate modules to get primary input embeddings. Second, each of these embeddings is fed to a Transformer encoder \cite{vaswani2017attention} followed by a max pooling layer, which yields the latent embeddings. Third, the latent embeddings are fed to the alignment and heterogeneity modules to generate the embeddings $H^{align}$ and $H^{het}$. Last, we concatenate $H^{align}$, $H^{het}$ and the debate meta-data $X_M$ which is fed to an MLP for persuasiveness prediction. The latent embeddings interact with two procedures alternately: optimize the alignment loss $L_{align}$ and persuasiveness loss $L_{pers}$, and learn weights through 3 reference models.}
	\label{fig:archi}
\end{figure*}

\nop{
The principal objective of this paper is to predict the persuasiveness of a given speech from a debate video clip. More specifically, we are interested in predicting, given the number of votes before a debater speaks, how many votes the debater will get after his/her speech (e.g., 1 minute) using  acoustic, visual, linguistic and debate meta-data.} Figure~\ref{fig:archi} shows an overview of our \PPAMI\ architecture with a brief description of its major components. Note that the key novelties of this paper are the two novel modules (i.e., the alignment module and the heterogeneity module shaded in yellow in Figure~\ref{fig:archi}) that constitute the adaptive fusion framework (Section~\ref{sec:adaptive_fusion}) \footnote{Our proposed adaptive fusion framework has the potential of being broadly utilized in other multimodal learning tasks. We leave that exploration for future work.}.

\subsection{Generating Primary Input Embeddings}

Given a video clip, we respectively represent the acoustic, visual and language input as $X_A\in \mathcal{R}^{T_A}$, $X_V\in \mathcal{R}^{(H\times W\times C)\times T_V}$, $X_L\in \mathcal{R}^{D\times T_L}$, where $T_A, T_V, T_L$ are respectively the lengths of the audio signal, face sequence, and word sequence. $H,W,C$ are the height, width and the number of channels of each image, and $D$ is the length of our dictionary of words. In addition, we also use two debate meta-data features: the number of votes before a speech and the length of the speech. We generically denote the debate meta-data as a vector $X_M\in\mathcal{R}^{d_M \time 1}$, where $d_M=2$.

We first extract features from the three modalities, then add a fully-connected (FC) layer for each modality to obtain low dimensional primary input embeddings. The generated primary input embeddings are depicted as multi-dimensional bars (as a symbol of vector sequences) in Figure~\ref{fig:archi}. Here we describe the detailed feature extraction components.

\noindent\textbf{Feature extraction from the acoustic modality.} For each audio clip, we use Covarep \cite{degottex2014covarep} to extract MFCCs\footnote{The energy-related 0th coefficient is excluded}, Glottal source parameters, pitch-related features, and features using the Summation of Residual Harmonics method \cite{drugman2011joint}. These features capture human voice characteristics from different perspectives and are all shown to be relevant to emotions \cite{ghosh2016representation}. These 73 dimensional features are averaged over every half second.

\noindent\textbf{Feature extraction from the visual modality.} Since the speakers in both datasets can be highly dynamic and occluded, we capture only their faces as Brilman et al.~\cite{brilman2015multimodal} did to reduce noisy input. The details of face detection and recognition are in Section~\ref{sec:dataset}. Given each facial image, we use the VGG19 architecture~\cite{simonyan2014very} pre-trained on the Facial Emotion Recognition FER2013 dataset\footnote{https://www.kaggle.com/c/challenges-in-representation-learning-facial-expression-recognition-challenge/overview} and extract the 512 dimensional output before the last FC layer as the face features.

\noindent\textbf{Feature extraction from the language modality.} We use the Jieba\footnote{https://github.com/fxsjy/jieba} Chinese text segmentation library to segment Chinese sentences (utterances) into words. We use the Tencent Chinese embedding corpus \cite{song-etal-2018-directional} to extract 200 dimensional word embeddings. In the case of English, we extract 64 dimensional Glove word embeddings \cite{pennington2014glove} trained from all transcripts from the IQ2US debates.

All features are passed to a learnable FC+ReLU layer which converts the initial features into \emph{primary input embeddings}. The primary input embeddings thus obtained for each of the three modalities are respectively $H_A^{in}\in \mathcal{R}^{d_{in}\times T_A'},H_V^{in}\in \mathcal{R}^{d_{in}\times T_V'},H_L^{in}\in \mathcal{R}^{d_{in}\times T_L'}$, where $d_{in}=16$ is the row-dimension of the primary input embeddings, which is same across different modalities. $T_A', T_V', T_L'$ denote the sequence lengths of the modalities, where $T_V'=T_V, T_L'=T_L$.
Note that in our primary input embeddings, the timestamps of the acoustic, visual, and language modality respectively represent a short time window, a frame, and a word. 

\subsection{Generating Compact Latent Embeddings of Modalities with Transformers}
To get a compact representation of the primary input embeddings for each modality, we aggregate the sequence of features into a single representation vector using one Transformer encoder per modality. Transformer encoders have been shown to outperform many other deep architectures, including RNNs, GRUs, and LSTMs in many sequential data processing tasks in computer vision~\cite{wang2018non} and natural language processing~\cite{devlin2018bert}. The multi-head self-attention mechanism of Transformer better memorizes the long-term temporal dynamics~\cite{vaswani2017attention}.

With the Transformer encoder, the primary input embedding 
$H_m^{in}, m\in \{A,V,L\}$ of each modality is respectively transformed into a representation as:
\begin{align}
    H_m^{trans}=\text{TransformerEncoder}(H_m^{in}),
\end{align} where $H_m^{trans}\in\mathcal{R}^{d_{trans}\times T_m'}$, and $d_{trans} = 16$ is the dimension of the latent space after the Transformer encoder. 

To convert arbitrary length time sequences into standardized latent embedding vectors $H_{m}^{latent}\in\mathcal{R}^{d_{trans}\times 1}$, we additionally use a max pooling layer:
\begin{align}
    H_m^{latent}=\text{MaxPool}(H_m^{trans}).
\end{align}
$H_m^{latent}$ intuitively captures the maximum activation over the time sequence along each dimension of $d_{trans}$.

\subsection{Balancing Shared and Heterogeneous Information with Adaptive Fusion}\label{sec:adaptive_fusion}
As mentioned earlier, there are two conflicting aspects of multimodal data. First, data from different modalities within the same time frames may sometimes be highly aligned (i.e., have shared information). Second, different modalities may sometimes contain diverse cues which may not be equally important for prediction. To balance the aligned and heterogeneous multimodal information, we propose a novel adaptive fusion framework consisting of two key modules: an \emph{alignment} module and a \emph{heterogeneity} module (shaded in yellow in Figure~\ref{fig:archi}).

\subsubsection{Alignment Module}

To extract information shared across different modalities, we first use a \emph{shared} multi-layer perceptron (MLP$^s$) to project the latent embeddings of each modality $m=A,V,L$ into the same latent space:
\begin{align}
    H_m^{s} = \text{MLP}^s(H_m^{latent})~
\end{align}
Here, $H_m^s \in \mathcal{R}^{d_s}$, where $d_s = 16$ is the dimension of the shared projection space. MLP$^s$ is shown as three rounded grey boxes in Figure~\ref{fig:archi}.

\edit{Inspired by existing multimodal representation learning work~\cite{andrew2013deep,dumpala2019audio} and the success of domain adaptation techniques~\cite{sun2016deep,het-da,mdd-da,localpreserve-da}, for each pair of modalities $\{m,n\}$, we use a cosine loss term $l_{cos}(m,n)$ and a domain adaptation loss term $l_{da}(m,n)$ to measure the alignment of $m,n$ in the shared projection space:  
\begin{equation}\label{eq:align-loss}
\mathcal{L}_{align} = \sum_{\{m,n\} \subset \{A,V,L\}} l_{cos}(m,n) + l_{da}(m,n)~
\end{equation}
The cosine loss term measures the similarity between embeddings of modalities $m,n$ of one sample:
\begin{equation}
l_{cos}(m,n) = 1 - cos(H_m^s, H_n^s)~\notag
\end{equation}
and the domain adaptation loss is one deep coral loss \cite{sun2016deep}, which measures the distance between the correlation matrices of modalities $m,n$:
\begin{equation}
l_{da}(m,n) = \frac{1}{4d_s^2}\norm{C^s_m - C^s_n}_F^2~\notag
\end{equation}
where $C_m$ and $C_n$ are correlation matrices of the embeddings in the shared space for modalities $m,n$ respectively. The cosine loss aligns the modalities in sample basis whereas the domain adaptation loss aligns the two modalities in a distribution level.
}


During training, the alignment loss $\mathcal{L}_{align}$ will be added to the entire prediction loss function as a regularization term to penalize lack of alignment between the 3 modalities in the projected space.

After the shared MLP layer, the regularized embeddings $H_m^s$ are in the same latent space. We apply mean pooling to average the three embeddings:
\begin{align}
    H^{align}= \text{MeanPool}(H_A^{s}, H_V^{s}, H_L^{s})~,
\end{align}
$H^{align} \in \mathcal{R}^{d_s}$ now contains shared information from all modalities.

\subsubsection{Heterogeneity Module}
Another key observation discussed in Section~\ref{sec:intro} is that different modalities may contain diverse information, and therefore make unequal contributions to the final prediction of persuasiveness (e.g., due to the noisy data from certain modalities as shown in Figure~\ref{fig:noisy-modality}).
We therefore propose a novel heterogeneity module which utilizes an interactive training procedure (Algorithm~\ref{alg:mirl}) to learn weights for different modalities. 

Intuitively, the importance of each modality should be inversely proportional to the ``error'' caused by the modality. To estimate this error term, we create three unimodal MLP reference models (represented as dashed arrows and rounded grey boxes at the central bottom of Figure~\ref{fig:archi}) parameterized by $\phi_A,\phi_V, \phi_L$ for the acoustic, visual, and language modalities respectively. Each unimodal MLP takes the compact latent embedding $H_m^{latent}$ generated by the Transformer encoder as input and generates a unimodal prediction $\hat{Y}_m^{ref}$ for each modality $m=A,V,L$:
\begin{align}
    \hat{Y}_m^{ref} = \text{MLP}_m^{ref}(\phi_m; H_m^{latent})~.
\end{align}
We use $T_{val}$ to denote the validation set, $Y_{val} \in \mathcal{R}^{|T_{val}|}$ are the labels, and $\hat{Y}_{m,val}^{ref} \in \mathcal{R}^{|T_{val}|}$ are the predictions made by the unimodal reference model for modality $m$.
The reference models ($\phi_m$'s) are updated 
using the following Mean Squared Error (MSE) loss alone:
\begin{equation}\label{eq:ref_loss}
    \mathcal{L}_m^{ref}=\frac{ \norm{Y_{val} - \hat{Y}_{m,val}^{ref}}_2^2  }{ |T_{val}| }
\end{equation} 

After several epochs of training $\phi_m$'s, we are able to obtain a converged MSE loss of each reference model. We then use the updated reference model to estimate the prediction errors by $\mathcal{L}_m^{ref}$.
$\mathcal{L}_m^{ref}$ is used to guide the weights $w_m$ of latent embeddings $H_m^{latent}$ ($m = A,V,L$) to be concatenated in the heterogeneity module:
\begin{align}\label{eq:weighted_concat}
    H^{het}=w_AH_A^{latent}\oplus w_VH_V^{latent} \oplus w_LH_L^{latent}.
\end{align} 
$w_A,w_V,w_L$ are scalars incrementally updated over epochs:
\begin{align}\label{eq:update_concat_weight}
    w_m = \alpha w_m + (1-\alpha) \tilde{w}_m,
\end{align} 
where $\alpha\in (0,1)$ controls the rate of update, and $\tilde{w}_m$ is obtained using the following softmax function of the reference model validation losses:
\begin{align}\label{eq:get_concat_weight}
\tilde{w}_m = \frac{\exp\{-\beta \mathcal{L}_m^{ref}\}}{\sum_{m'=A,V,L}\exp\{-\beta \mathcal{L}_{m'}^{ref}\}},\forall m=A,V,L
\end{align} $\beta>0$ is a scaling factor. Since $\sum_m \tilde{w}_m=1$, combining Equation~\eqref{eq:update_concat_weight}, it is guaranteed that $\sum_m w_m=1$. 

\subsubsection{Adaptive Fusion with Interactive Training}
The representations obtained from the alignment module ($H^{align}$)
and the heterogeneity module ($H^{het}$) are then concatenated together with the debate meta-data $X_M$ and fed into a final MLP layer to make the final prediction $\hat{Y}$:
\begin{align}
    \hat{Y} = f(\theta;\!X_{A},X_{V},X_{L},X_{M}) = \text{MLP}(H^{align}\oplus H^{het}\oplus X_M)
\end{align}
where $\theta$ is the set of parameters of the \PPAMI\ model excluding the reference model parameters $\phi_m$. \edit{The modality weights in $H^{het}$ are adapted from the losses of the unimodal models. Together with $H^{align}$, the representations from both modules are learned adaptively through an interactive training process.}

To train the \PPAMI\ model, we have two loss terms: a novel alignment loss $\mathcal{L}_{align}$, and a persuasiveness loss term $\mathcal{L}_{pers}$. In the case of the IPP problem, $\mathcal{L}_{pers}$ is the MSE loss. In the case of DOP, we use  cross-entropy loss for the binary classification.
The total loss function is a weighted combination:
\begin{align}\label{eq:final_loss}
  \mathcal{L}_{final}=  \mathcal{L}_{pers}+\gamma \mathcal{L}_{align},
\end{align} where $\gamma$ is a weight factor.

The entire training proceeds in a master-slave manner, as shown in Algorithm~\ref{alg:mirl}. In each epoch of the master training procedure (Lines 4 to 14), we use the total loss function in Equation~\eqref{eq:final_loss} to update the parameters $\theta$ of the main \PPAMI\ components. The  weights $w_A,w_V,w_L$ of the 3 modalities are obtained using reference models $\phi_m$, and their losses $\mathcal{L}_m^{ref}$ are then updated in the slave procedure. In each epoch of the slave procedure (Lines 8 to 10), we take the latent embeddings from the master procedure as input and update the reference models with the loss function in Equation~\eqref{eq:ref_loss}. We then obtain the weights $w_A, w_V, w_L$ of different modalities in the heterogeneity module. 

\begin{algorithm}[ht]
\SetAlgoLined
\DontPrintSemicolon
\KwIn{Training dataset $T$, validation datset $T_{val}$; Number of epochs $n$ and $N$}
\KwOut{Multi-modality model $f(\theta;X_A,X_V,X_L,X_M)$, modality weights $w_m$ ($\forall m = A,V,L$)}

Initialize three unimodal reference models $\phi_m (\forall m=A,V,L)$ and $\theta$;

Initialize $w_A=w_V=w_L=1/3$;

\% Master Procedure Start

\For{epoch=1,\ldots,N}{
Update $\theta$ with loss function Equation~\eqref{eq:final_loss};

Get latent embeddings $H_m^{latent}, \forall m=A,V,L$;

  \% Slave Procedure Start
  
  \For{epoch=1,\ldots,n}{
  Update $\phi_m, \forall m=A,V,L$ with loss function in Equation~\eqref{eq:ref_loss};
  }\% Slave Procedure End

  Get reference model losses $\mathcal{L}_m^{ref},\forall m=A,V,L$;
  
  Update modality importance weights $w_m,\forall m=A, V, L$ using Equations~\eqref{eq:update_concat_weight}-\eqref{eq:get_concat_weight};

}\% Master Procedure End

\Return $\theta$, $w_m (\forall m = A,V,L)$

\caption{\PPAMI\ interactive training procedure}\label{alg:mirl}
\end{algorithm}

\section{Datasets}\label{sec:dataset}
We describe our two datasets below. 
\vspace{-1mm}
\subsection{QPS Dataset}
\emph{We created the QPS dataset by getting videos\footnote{An example can be found in https://youtu.be/P5ehhs0hpFI.} from the popular Chinese TV debate show Qipashuo.} In each episode of the TV show, 100 audience members initially vote `for' or `against' a given debate topic. Debaters from `for' and `against' teams speak alternately, and the audience can change their votes anytime. In general, there are 6--10 speech turns. Final votes are turned in after the last speaker. The winner is the team which has more votes at the end than at the beginning. 
For example, if the initial and final `for' vs. `against' votes are 30:70 and 40:60, respectively, then the `for' team wins because they increased their votes from 30 to 40 (even though they still have fewer votes than the ``against'' team).
In total, we collected videos of 21 Qipashuo episodes with 205 speaking clips spanning a total of 582 minutes. 

 We extracted the transcripts from the video subtitles. To sufficiently preprocess the videos for subtitle extraction, we took the following steps. First, we sampled 2 frames per second and binarize the images with a threshold 0.6, which can avoid the influence from various colors of subtitles in videos. Second, we cropped the subtitles by a fixed bounding box since the position of subtitles is fixed in all the videos.
Third, we clustered the binarized images into buckets such that any two binarized images in the same bucket are identical on 90\% or more pixels. We then randomly selected one of these images to represent the cluster.
This helps reduce noise (e.g. from advertisements displayed on the image). Finally, the surviving binary images were fed into an OCR API to get accurate transcripts. We used Baidu's off-the-shelf pre-trained OCR API\footnote{https://ai.baidu.com/tech/ocr}, so no extra data is needed for training.

If we take each speaking clip as a train/test instance, there would be a total of 205 data points. This paucity of information poses a huge challenge for machine learning. We therefore segment each speaking clip into clips of 50 utterances each according to the transcript we extract above. Note that 50 is the smallest number of utterances in any speaking clip of our dataset. Moreover, note that these ``sub-clips'' of 50 utterances yield a temporal sequence whose temporal dynamics can be important.
The labels are shared for segments extracted from the same clip. This trick yields 2297 such segments which are used as train/test instances in our evaluation.

As the speakers are highly dynamic and often occluded, we only use speakers' faces as the visual input. We extract 2 frames per second from videos and use Dlib\footnote{http://dlib.net} for face detection and recognition. The recognition is based on one pre-annotated profile for each speaker and is only needed for training.
To further reduce false positives (i.e., extracting the face of the non-speakers), we first use the model from \cite{ijcai2019-626} to remove faces in the image that are not speaking, and then use the method from \cite{Marin19a} for face tracking.

\subsection{IQ2US}\label{sec:data-iq2us}
We also evaluate \PPAMI\ on the benchmark IQ2US TV debate dataset used by \cite{brilman2015multimodal,santos2016domain,zhang2016conversational,potash2017towards,wang2017winning}. This dataset was originally collected by~\cite{brilman2015multimodal}. The audience can only vote at the beginning and at the end of the debate, and the winner is determined in the same way as in Qipashuo. Note that we cannot use the same set of videos as \cite{brilman2015multimodal}, since they were interested in predicting the result of the whole debate, which doesn't require the transcripts to be aligned within shorter clips. Of the 100 episodes we collected, only 58 had transcripts that were correctly aligned with the visual modality at the minute level. Finally, we get 852 one-minute single-speaker clip instances from the 58 episodes --- 428 of them belong to the winning side. As transcripts are available in the IQ2US data, no pre-processing is required for the language modality in this dataset. For the visual modality, we use the same procedures as in the QPS dataset to extract the face image sequences of the speakers.
Since there are no intermediate votes in IQ2US,  we only predict the debate outcome (i.e. whether a single-speaker clip instance belongs to the winning team).

\begin{table*}[htp!]
\begin{center}
\begin{tabular}{|l|cccccccccc|c|}
\hline
Fold & 1&2&3&4&5&6&7&8&9&10&Average \\
\hline\hline
Brilman et al.  \cite{brilman2015multimodal} (early fusion) & 0.009&0.011& 0.016& 0.017& 0.030& 0.018& 0.020& 0.012& 0.013& 0.018& 0.016 \\
Nojavanasghari et al. \cite{nojavanasghari2016deep} (late fusion) & 0.007 & 0.015 & 0.019 & \textbf{0.011} & 0.027 & \textbf{0.014}& 0.020& 0.012& 0.020& 0.015&0.016 \\
Santos et al. \cite{santos2018multimodal} (early fusion) &0.025 &0.019 &0.018 &0.019 &0.018 &0.017 &0.029 &0.016 &0.024 &0.018 & 0.020\\
\editrow Verma et al. \cite{verma2019deepcu} & 0.012 & 0.013 & 0.016 & 0.012 & 0.021 & 0.016 & 0.016 & 0.019 & 0.012 & 0.016 & 0.015\\
\hline\hline
\PPAMI\ without DA loss & 0.006 & 0.010 & 0.015 & 0.015 & 0.020 & 0.015 & 0.012 & \textbf{0.009} & 0.009 & 0.013 & 0.012 \\
\editrow \textbf{\PPAMI\ (proposed method)} & \textbf{0.006}& \textbf{0.010}& \textbf{0.011}& 0.014& \textbf{0.017} & 0.015 & \textbf{0.012}& 0.010& \textbf{0.008}& \textbf{0.012} & \textbf{0.011}\\
\hline
\editrow $dec.$ \% & 14.2 & 9.1 & 31.3 & -27.3 & 5.6 & -7.1 & 25.0 & 16.7 & 33.3 & 20.0 & 26.7
\\
\hline
\end{tabular}
\end{center}

\caption{MSE for each test fold of different approaches to solving the Intensity of Persuasion Prediction (IPP) on the QPS Dataset. The last row shows the MSE decrease percentage of \PPAMI\ compared to the best baseline in each fold. \edit{DA loss stands for the domain adaptation loss $l_{da}$. On average, \PPAMI\ achieves a lower MSE than the baselines by at least 26.7\%, which is statistically significant with $\text{p-val}  < 0.05$. Note that the vote scores we predict are normalized.}}
\label{tab:qp-mirl}
\end{table*}

\begin{table}[h]
\begin{center}
\begin{tabular}{|l|c|c|}
\hline
Method & DOP (Accuracy)\\
\hline\hline
 Brilman et al. \cite{brilman2015multimodal} (early fusion) &  0.614\\
 Nojavanasghari et al. \cite{nojavanasghari2016deep} (late fusion)& 0.615\\
 Santos et al. \cite{santos2018multimodal} (early fusion)& 0.598\\
 \editrow Verma et al. \cite{verma2019deepcu} (DeepCU) & 0.622\\
 \hline\hline
 \PPAMI\ without DA loss & 0.635 \\
 \editrow \PPAMI\ with MDD DA loss & 0.629 \\
 \editrow \textbf{\PPAMI\ (proposed method)} & \textbf{0.639} \\
\hline
\end{tabular}
\end{center}

\caption{Prediction accuracy for Debate Outcome Prediction in IQ2US dataset. \edit{Our \PPAMI\ is 1.7\%--2.5\% better than baselines. The DA stands for domain adaptation and the MDD DA loss is employed from \cite{mdd-da}. \PPAMI\ improvements over baselines are statistically significant with $\text{p-val}  < 0.05$.}}
\label{tab:iq2-mirl}
\end{table}

\section{Experimental Evaluations}
Our experiments assess the performance of \PPAMI\ on the DOP and IPP tasks. Specifically:
\begin{enumerate}
\item \edit{(IPP) We predict the change of votes after a speech by a debater --- this is done on the QPS dataset. Note that the number of votes are scaled by the total number of audience members and hence is guaranteed to lie in the  [0,1] interval. Hence, the change of votes always lies in the [-1,1] interval.}
\item (DOP) We predict whether a clip in which a debater is speaking is part of the winning team of the debate --- this is done on the IQ2US dataset.
\end{enumerate}
In addition, we also conduct an ablation study that assesses the contributions of different components of \PPAMI. Moreover, we assess the importance of different modalities as well as time frames using the QPS dataset. \edit{Finally, we test the sensitivity of the optimal hyper-parameters used in the model.}.

\subsection{Experimental Settings}
QPS uses a 10-fold rolling window prediction. Specifically, we construct 10 sequences of consecutive episodes of the show. For instance, if $E_1,\ldots,E_k$ represent the set of all QPS episodes, then one sequence would be $Seq_{k}=E_1,\ldots, E_{k}$, another would be $Seq_{k-1}=E_1,\ldots, E_{k-1}$.  For any such sequence $Seq_i = E_1,\ldots, E_{i}$, we set $E_{i}$ as the test episode (i.e. the episode on which we make predictions). We learn a model from the first $i-3$ episodes $E_1,\ldots,E_{i-3}$ and identify the best parameters for our model by using episodes $E_{i-2},E_{i-1}$ as the validation set.  As the same subject can occur in multiple episodes of QPS, in order to avoid information leakage from training to test data, we do not train a model from $E_i$ to predict $E_{j,j<i}, \forall i,j$.

For IQ2US, 10-fold cross validation is used since a debater can only appear in one episode. The initial vote score and speaking length features are normalized to $[0,1]$.

Denote FC\textit{n} as a fully-connected layer that outputs \textit{n}-dimension vectors. The MLPs in the reference models and final multimodal prediction model are all configured as FC16+ReLU, FC8+ReLU, and FC1+Sigmoid. The shared MLP in \emph{alignment module} is FC16+ReLU.
\PPAMI\ uses Batch Normalization~\cite{ioffe2015batch} right after each of the FC layers for input embeddings, and uses 0.4 as dropout \cite{hinton2012improving} after all FC16 layers.
For the Transformer encoder, we use a single layer with 4 heads, where the input, hidden, and output dimension are all 16.
We use the Adam \cite{kingma2014adam} optimizer with a weight decay of $10^{-5}$. The numbers of epochs in Algorithm 1 is $N=200$ and $n=10$.
The learning rate $lr$, alignment loss weight $\gamma$, update scalar $\alpha$, scaling factor $\beta$ are finalized by grid search. We ended up using $lr=0.001, \gamma=0.1, \alpha=0.5, \beta=50$ as these yield the best results on the validation sets.

\subsection{Comparison with Baselines}
We compare both tasks with three multimodal persuasion prediction baselines: early fusion + SVM \cite{brilman2015multimodal}, deep multimodal late fusion \cite{nojavanasghari2016deep},  early fusion + LSTM \cite{santos2018multimodal}, \edit{and a more recent multimodal fusion baseline, DeepCU \cite{verma2019deepcu}}.
Brilman et al. \cite{brilman2015multimodal} extract audio, visual and linguistic features from IQ2US debate videos and concatenate these features, which are fed into an SVM for classification. Although \cite{brilman2015multimodal} also solves the DOP problem on the IQ2US dataset, it is different from our work in that (i) the used episodes are different (see Section \ref{sec:data-iq2us} and (ii) it uses long video inputs (9--36 minutes) of all debates while we only use a short speaking clip (1 minutes) of a single speaker. Thus, for fair comparison, we implemented their method and ran experiments in our data. Nojavanasghari et al. \cite{nojavanasghari2016deep} first feed features of each modality to a neural network to get predictions of the modality, then uses a fusion neural network to combine the modality-based predictions. Santos et al. \cite{santos2018multimodal} model the temporal dynamics by using an LSTM on the concatenated features from all modalities. \edit{Verma et al. \cite{verma2019deepcu} propose a deep model to integrate both common and unique latent information for multimodal sentiment analysis. }

In the case of the IPP problem, we adapt the first baseline by modifying it to use an SVM regressor (rather than an SVM classifier). For the other three baselines, we replace the final layers by a regression and use MSE loss to train the models. For fairness, we also allow the baselines to use the two debate meta-data features. 
The results comparing \PPAMI\ on IPP and DOP with past approaches are shown in Tables \ref{tab:qp-mirl} and \ref{tab:iq2-mirl}, respectively. 

\emph{IPP Problem.}
Table \ref{tab:qp-mirl} shows the MSE obtained by different approaches in each fold and the average on the QPS dataset. Note that the vote scores are normalized to lie in the $[0,1]$ interval. The last line of Table~\ref{tab:qp-mirl} shows the decrease percentage of MSE which is defined as $dec.$ = 1-MSE(\PPAMI)/MSE(the best baseline).
For instance, from the first column of Table~\ref{tab:qp-mirl}, we see that the percentage decrease is $1-\frac{0.006}{0.007}\approx 0.14$ representing a 14\% decrease of MSE generated by \PPAMI\ compared to the best of the baselines.
\edit{We observe that on average, \PPAMI\ yields a 26.7\% decrease of MSE compared with the best baseline which is statistically significant via a Student t-test ($\text{p-val}  < 0.01$). In addition, the comparison of the last two methods shows that the domain adaptation loss $l_{da}$ in Equation \ref{eq:align-loss} improves the performance by $1 -\frac{0.011}{0.012}=8.3\%$.} Moreover, \PPAMI\ is more robust and performs better than all baselines in 7 out of 10 folds. 

\emph{DOP Problem.} Table \ref{tab:iq2-mirl} shows the average prediction accuracy over 10 folds on the DOP problem w.r.t. the IQ2US dataset. \edit{When we compare \PPAMI\ (last row) with the four baselines, it is clear that \PPAMI\ achieves 1.7\%--2.5\% higher average accuracy than the baselines. The improvement is statistically significant ($\text{p-val} < 0.05$). To further investigate the domain adaptation (DA) loss ($l_{da}$ in Equation \ref{eq:align-loss}, we evaluate two variations of \PPAMI: \PPAMI\ without DA loss, and \PPAMI\ with the MDD DA loss \cite{mdd-da}. We observe that the proposed method (with deep coral DA loss) achieves the best, although the difference among the three are \emph{not} statistically significant (p-val = 0.08).}

Overall, the two experiments make \PPAMI\ the best performing system for both the IPP and the DOP problems.

\begin{table}[h]
\begin{center}
\begin{tabular}{|l|c|}
\hline
Method & MSE\\
\hline\hline
\editrow \PPAMI\ without DA loss ($l_{da}$) & 0.012 \\ 
 \editrow \PPAMI\ without alignment loss& 0.018 \\
 \editrow \PPAMI\  without reference models&  0.014\\
 \editrow \PPAMI-LSTM&  0.033\\
 \editrow \PPAMI-Acoustic (unimodal) &  0.017\\
 \editrow \PPAMI-Visual (unimodal) & 0.019 \\
 \editrow \PPAMI-Language (unimodal) & 0.016\\
 \hline
 \editrow \textbf{\PPAMI}  & \textbf{0.011}\\
\hline
\end{tabular}
\end{center}

\caption{Ablation study results. All improvements are statistically significant ($\text{p-val}  < 0.05$).}
\label{tab:ablation}
\end{table}

\subsection{Ablation Study}
To measure the contributions of the different components of \PPAMI, we create four methods, each with one component removed from \PPAMI\ :
\begin{itemize}
    \item \edit{\PPAMI\ without the domain adaptation (DA) loss $l_{da}$}.
    \item \edit{\PPAMI\ without the alignment loss ($l_{da}$ and $l_{cos}$).}
    \item \PPAMI\ without reference models. The latent embeddings are concatenated by equal weights $1/3$.
    \item \PPAMI-LSTM. The Transformer encoder and max pooling layer are replaced by a 1-layer LSTM.
    \item \PPAMI-unimodal. We input a single modality without alignment loss and latent embedding concatenation. That is, the latent embedding is directly concatenated with the debate meta-feature and fed to the final MLP.
\end{itemize}

\edit{Table~\ref{tab:ablation} shows the average MSE obtained on the QPS dataset for both \PPAMI\ and the 7 variations above. First,  rows 2,3 and the last row show that if \PPAMI\ does not use the alignment module and reference models in the heterogeneity module, the MSE increases from 0.011 to 0.018 and 0.014 respectively. This is statistically significant ($\text{p-val} < 0.05$) and hence shows the power of both proposed adaptive fusion modules in Section~\ref{sec:adaptive_fusion}}. Second, we observe the power of the Multihead-attention Transformer encoder to handle long sequences, as the \PPAMI-LSTM model achieves the worst MSE amongst all methods. Third, we observe from rows 4-6 that the language modality is the most important in the prediction task, while the acoustic and visual modalities are less important. 

\subsection{Visualization of Prediction}
In this experiment, we show (1) the importance of modalities through their learned weights (cf. Equation~\eqref{eq:weighted_concat}), and (2) the examples of learned temporal attention weights from different modalities.

\begin{figure}[t]
	\centering
		\includegraphics[width=\columnwidth,scale=0.6]{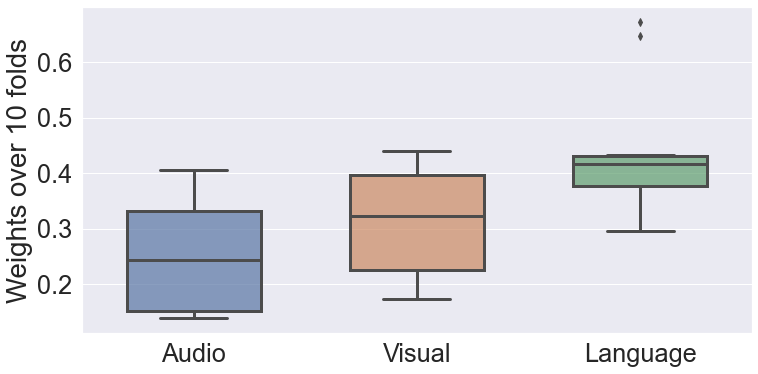}
\caption{Modality weights in the heterogeneity module.}
	\label{fig:guide-weights}

\end{figure}

\noindent\textbf{Modality weights.} We report the modality weights in the heterogeneity module of the trained \PPAMI\ in all folds of QPS dataset. Figure~\ref{fig:guide-weights} shows box plots for the three modalities. \edit{The language modality is the most important and robust over all folds with a median weight of 0.42, while the median weights of acoustic and visual modalities are $0.24$ and $0.33$ respectively.}

\begin{figure}[t]
	\centering
		\includegraphics[width=\columnwidth,scale=1]{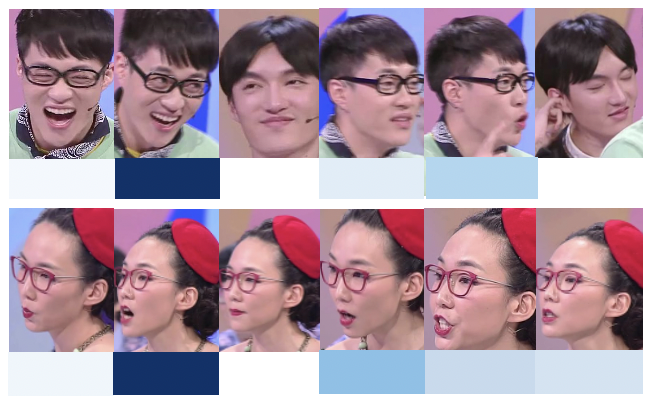}
\caption{Temporal attention of visual modality -- color coded as blue. Darker color implies higher attention weight.}
	\label{fig:vis-mod-attn}
	
\end{figure}

\begin{figure}[t]
	\centering
		\includegraphics[width=\columnwidth,scale=1]{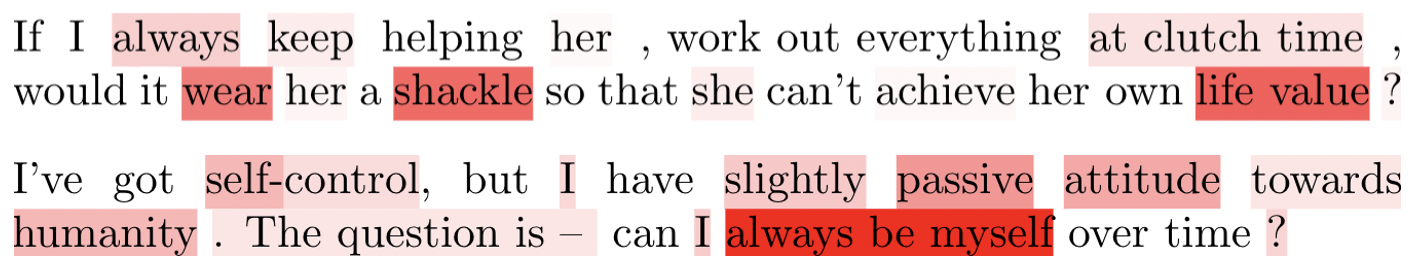}
\caption{Temporal attention of language modality -- color coded as red. Darker color implies higher attention weight. The original Chinese transcripts are translated to English.}

	\label{fig:lang-mod-attn}
\end{figure}

\noindent\textbf{Temporal attention weights.}
We visualize the temporal attention weights of two sample sequences of visual (Figure~\ref{fig:vis-mod-attn}) and language (Figure~\ref{fig:lang-mod-attn}) modalities. 
\edit{For each timestamp $t$, assume $\alpha_{i,t}^{(h)}$ is the scaled dot-product attention weight from query $i$ to time $t$ in head $h$,  learned by the Transformer Encoder \cite{vaswani2017attention}. We calculate the temporal attention weight at time $t$ as 
$$a_t = \frac{1}{IH}\sum_{i=1}^I\sum_{h=1}^{H}\alpha_{i,t}^{(h)},$$
which represents the amount of attention the model pays to time $t$, where $I$ and $H$ are the number of queries and heads.}
In Figure~\ref{fig:vis-mod-attn} (top), the man's face is not detected correctly in frames 3 and 6 -- and we see that \PPAMI\ assigns near-zero attention weights to both frames, suggesting that these frames should be ignored. Moreover, the happy expression in frame 2 gets a high attention weight. The woman below gets high attention weights when she actively talks to someone (frames 2,4,5). In Figure~\ref{fig:lang-mod-attn}, we notice that reasonable keywords like `wear', `shackle', `passive', and `hold' also get high attention weights. Therefore, our \PPAMI\ captures the meaningful long-range temporal dynamics with the help of Transformer Encoder.

\edit{\subsection{Parameter sensitivity analysis}
We study the sensitivity of the hyper-parameters around their optimal values obtained from grid search. The update rate $\alpha$ of modality weights in Equation \ref{eq:update_concat_weight}, scalar $\beta$ for Softmax in Equation \ref{eq:get_concat_weight}, and weight $\gamma$ of the alignment loss in Equation \ref{eq:final_loss} are studied. We perturb each of them by $\pm5\%$ while fixing the other two. We use the modified hyper-parameter value and evaluate the relative change to the original accuracy 0.639 obtained by \PPAMI\ on DOP task (Table \ref{tab:iq2-mirl}). As a result, the relative change is at most 1.1\%, 0.8\% and 1.7\% for perturbing $\alpha, \beta, \gamma$ respectively. $\alpha$ is the most sensitive among the three, implying that the alignment loss is important in training the model. However, all the changes are within a very small range, which indicates that our model is robust.
}

\section{Discussion}
\subsection{Text Encoder Comparison for Linguistic Inputs}
In \PPAMI\ , the sequence of word embeddings is used as the sequence input to the Transformer encoder. Another way is to encode each sentence to an embedding and feed the sequence of sentences to the Transformer encoder. We have conducted experiments to compare these two methods. To get English sentence embeddings in IQ2US, we employ the pre-trained Universal Sentence Encoder \cite{cer2018universal} in TFHub\footnote{https://tfhub.dev/google/universal-sentence-encoder/1}. For Chinese sentences in QPS, we train an LSTM to get 128-dimensional sentence embeddings. 
We replace word embeddings with sentence embeddings and conduct the experiments in both datasets. As a result, the accuracy in IQ2US is 0.623 (1.4\% worse than \PPAMI) and the mean squared error in QPS is 0.014 (20\% worse than \PPAMI). Thus, the fine-grained word-level embeddings are better than sentence-level embeddings. The word order and semantic meaning is already captured by the word-level embeddings.

\subsection{Heterogeneity Module \textit{vs.} Attention Mechanism}
Intuitively, the heterogeneity module in \PPAMI\ aims to learn the modality-wise importance from data. An alternative is to use the attention mechanism to attend the model to different modalities. However, the attention mechanism introduces extra amount of trainable parameters into \PPAMI\ . In our early experiments, this resulted in worse results due to the small dataset (2297 and 805 data points for QPS and IQ2US resp.). On the contrary, the parameters introduced by heterogeneity module are independent from the rest of \PPAMI\ model, which fuses modalities and achieves better prediction results.

\edit{\subsection{Modality importance} The  importance of the different modalities is shown in two experiments: ablation study (Table \ref{tab:ablation}) and modality weights (Figure \ref{fig:guide-weights}). Table \ref{tab:ablation} shows that when using a single modality as input, the language modality is the best while the visual modality is the worst. Yet when the three modalities are combined together in \PPAMI, Figure \ref{fig:guide-weights} shows that the audio modality makes the smallest contribution while the language modality is still the strongest. }

\edit{
\section{Limitations and Future Work}
}

First, it is important to note that the adaptive fusion technique in \PPAMI\ can be generalized to other multimodal sequence prediction problems such as video question answering and video sentiment analysis. We leave this exploration for future work.

\edit{Second, since the importance of modalities can vary from sample to sample, a future effort could  investigate methods to learn sample-specific modality weights (e.g. different attention mechanisms), which might further improve the performance.} 

\edit{Third, prior knowledge is an important factor for humans when judging whether a speaker is persuasive or not. We were unable to assess the impact of prior knowledge in this work as we used publicly available datasets from TV shows. An alternative mechanism would be to conduct such debates ourselves with a studio audience who fill out a pre-debate survey that sheds light on their prior knowledge and then gets their votes periodically during and after the debate. This could be an important experiment to conduct under appropriate IRB protocols.
}

\edit{Last, in real-world persuasion challenges,  factors such as logic and the structure of arguments may be more important than the TV shows we have studied where acting and theatrics may be unreasonably important. An important future effort might run IRB-approved experiments involving such persuasion challenges data and use that to predict persuasion in other settings.}

\section{Conclusion}
In this paper, we have solved two problems. First, we provide a solution to the Debate Outcome Prediction (DOP) problem that improves on past work by 2\%--3.7\%. Though these numbers are not huge, they are statistically significant. Second, we are the first to pose and solve the Intensity of Persuasion Prediction (IPP) problem. We show that we are able to beat baselines built on top of past solutions to IPP by 25\% on average.  
Our proposed \PPAMI\ framework leverages both the common and modality-specific information contained in multimodal sequence data (audio, video, language), while learning to focus attention on the meaningful part of the data.  Moreover, our newly created QPS dataset provides a valuable new asset for future research --- it will be released upon publication of this paper.

\section*{Acknowledgment}
We gratefully acknowledge the support of NSF under Nos. OAC-1835598 (CINES), OAC-1934578 (HDR), CCF-1918940 (Expeditions), IIS-2030477 (RAPID), IIS-2027689 (RAPID);
DARPA under No. N660011924033 (MCS);
ARO under Nos. W911NF-16-1-0342 (MURI), W911NF-16-1-0171 (DURIP);
Stanford Data Science Initiative, Wu Tsai Neurosciences Institute, Chan Zuckerberg Biohub, Amazon, JPMorgan Chase, Docomo, Hitachi, JD.com, KDDI, NVIDIA, Dell, Toshiba, UnitedHealth Group, Adobe, Facebook, Microsoft, and IDEaS Insitute. J. L. is a Chan Zuckerberg Biohub investigator. Special thanks to Xinyu Cong for annotating the QPS dataset and Yuxuan Zhang for translating the Chinese transcripts in the figures of this paper.

\ifCLASSOPTIONcaptionsoff
  \newpage
\fi

\bibliographystyle{IEEEtran}
\bibliography{ref}

\begin{thebibliography}{10}
\providecommand{\url}[1]{#1}
\csname url@samestyle\endcsname
\providecommand{\newblock}{\relax}
\providecommand{\bibinfo}[2]{#2}
\providecommand{\BIBentrySTDinterwordspacing}{\spaceskip=0pt\relax}
\providecommand{\BIBentryALTinterwordstretchfactor}{4}
\providecommand{\BIBentryALTinterwordspacing}{\spaceskip=\fontdimen2\font plus
\BIBentryALTinterwordstretchfactor\fontdimen3\font minus
  \fontdimen4\font\relax}
\providecommand{\BIBforeignlanguage}[2]{{%
\expandafter\ifx\csname l@#1\endcsname\relax
\typeout{** WARNING: IEEEtran.bst: No hyphenation pattern has been}%
\typeout{** loaded for the language `#1'. Using the pattern for}%
\typeout{** the default language instead.}%
\else
\language=\csname l@#1\endcsname
\fi
#2}}
\providecommand{\BIBdecl}{\relax}
\BIBdecl

\bibitem{brilman2015multimodal}
M.~Brilman and S.~Scherer, ``A multimodal predictive model of successful
  debaters or how i learned to sway votes,'' in \emph{Proceedings of the 23rd
  ACM international conference on Multimedia}.\hskip 1em plus 0.5em minus
  0.4em\relax ACM, 2015, pp. 149--158.

\bibitem{nojavanasghari2016deep}
B.~Nojavanasghari, D.~Gopinath, J.~Koushik, T.~Baltru{\v{s}}aitis, and L.-P.
  Morency, ``Deep multimodal fusion for persuasiveness prediction,'' in
  \emph{Proceedings of the 18th ACM International Conference on Multimodal
  Interaction}.\hskip 1em plus 0.5em minus 0.4em\relax ACM, 2016, pp. 284--288.

\bibitem{santos2018multimodal}
P.~B. Santos and I.~Gurevych, ``Multimodal prediction of the audience's
  impression in political debates,'' in \emph{Proceedings of the 20th
  International Conference on Multimodal Interaction: Adjunct}.\hskip 1em plus
  0.5em minus 0.4em\relax ACM, 2018, p.~6.

\bibitem{vaswani2017attention}
A.~Vaswani, N.~Shazeer, N.~Parmar, J.~Uszkoreit, L.~Jones, A.~N. Gomez,
  L.~Kaiser, and I.~Polosukhin, ``Attention is all you need,'' in
  \emph{Advances in neural information processing systems}, 2017, pp.
  5998--6008.

\bibitem{verma2019deepcu}
S.~Verma, C.~Wang, L.~Zhu, and W.~Liu, ``Deepcu: integrating both common and
  unique latent information for multimodal sentiment analysis,'' in
  \emph{Proceedings of the 28th International Joint Conference on Artificial
  Intelligence}.\hskip 1em plus 0.5em minus 0.4em\relax AAAI Press, 2019, pp.
  3627--3634.

\bibitem{zhang2016conversational}
J.~Zhang, R.~Kumar, S.~Ravi, and C.~Danescu-Niculescu-Mizil, ``Conversational
  flow in oxford-style debates,'' in \emph{Proceedings of the 2016 Conference
  of the North American Chapter of the Association for Computational
  Linguistics: Human Language Technologies}, 2016, pp. 136--141.

\bibitem{potash2017towards}
P.~Potash and A.~Rumshisky, ``Towards debate automation: a recurrent model for
  predicting debate winners,'' in \emph{Proceedings of the 2017 Conference on
  Empirical Methods in Natural Language Processing}, 2017, pp. 2465--2475.

\bibitem{wang2017winning}
L.~Wang, N.~Beauchamp, S.~Shugars, and K.~Qin, ``Winning on the merits: The
  joint effects of content and style on debate outcomes,'' \emph{Transactions
  of the Association for Computational Linguistics}, vol.~5, pp. 219--232,
  2017.

\bibitem{habernal2016argument}
I.~Habernal and I.~Gurevych, ``Which argument is more convincing? analyzing and
  predicting convincingness of web arguments using bidirectional lstm,'' in
  \emph{Proceedings of the 54th Annual Meeting of the Association for
  Computational Linguistics (Volume 1: Long Papers)}, 2016, pp. 1589--1599.

\bibitem{joo2014visual}
J.~Joo, W.~Li, F.~F. Steen, and S.-C. Zhu, ``Visual persuasion: Inferring
  communicative intents of images,'' in \emph{Proceedings of the IEEE
  conference on computer vision and pattern recognition}, 2014, pp. 216--223.

\bibitem{huang2016inferring}
X.~Huang and A.~Kovashka, ``Inferring visual persuasion via body language,
  setting, and deep features,'' in \emph{Proceedings of the IEEE Conference on
  Computer Vision and Pattern Recognition Workshops}, 2016, pp. 73--79.

\bibitem{santos2016domain}
P.~B. Santos, L.~Beinborn, and I.~Gurevych, ``A domain-agnostic approach for
  opinion prediction on speech,'' in \emph{Proceedings of the Workshop on
  Computational Modeling of People’s Opinions, Personality, and Emotions in
  Social Media (PEOPLES)}, 2016, pp. 163--172.

\bibitem{andrew2013deep}
G.~Andrew, R.~Arora, J.~Bilmes, and K.~Livescu, ``Deep canonical correlation
  analysis,'' in \emph{International conference on machine learning}, 2013, pp.
  1247--1255.

\bibitem{9127152}
G.~{Song}, S.~{Wang}, Q.~{Huang}, and Q.~{Tian}, ``Learning feature
  representation and partial correlation for multimodal multi-label data,''
  \emph{IEEE Transactions on Multimedia}, pp. 1--1, 2020.

\bibitem{8489981}
N.~E.~D. {Elmadany}, Y.~{He}, and L.~{Guan}, ``Multimodal learning for human
  action recognition via bimodal/multimodal hybrid centroid canonical
  correlation analysis,'' \emph{IEEE Transactions on Multimedia}, vol.~21,
  no.~5, pp. 1317--1331, 2019.

\bibitem{aguilar-etal-2019-multimodal}
\BIBentryALTinterwordspacing
G.~Aguilar, V.~Rozgic, W.~Wang, and C.~Wang, ``Multimodal and multi-view models
  for emotion recognition,'' in \emph{Proceedings of the 57th Annual Meeting of
  the Association for Computational Linguistics}.\hskip 1em plus 0.5em minus
  0.4em\relax Florence, Italy: Association for Computational Linguistics, Jul.
  2019, pp. 991--1002. [Online]. Available:
  \url{https://www.aclweb.org/anthology/P19-1095}
\BIBentrySTDinterwordspacing

\bibitem{7984828}
L.~{Gao}, Z.~{Guo}, H.~{Zhang}, X.~{Xu}, and H.~T. {Shen}, ``Video captioning
  with attention-based lstm and semantic consistency,'' \emph{IEEE Transactions
  on Multimedia}, vol.~19, no.~9, pp. 2045--2055, 2017.

\bibitem{panagakis2015robust}
Y.~Panagakis, M.~A. Nicolaou, S.~Zafeiriou, and M.~Pantic, ``Robust correlated
  and individual component analysis,'' \emph{IEEE transactions on pattern
  analysis and machine intelligence}, vol.~38, no.~8, pp. 1665--1678, 2015.

\bibitem{7952687}
J.~{Pu}, Y.~{Panagakis}, S.~{Petridis}, and M.~{Pantic}, ``Audio-visual object
  localization and separation using low-rank and sparsity,'' in \emph{2017 IEEE
  International Conference on Acoustics, Speech and Signal Processing
  (ICASSP)}, 2017, pp. 2901--2905.

\bibitem{jo2019cross}
D.~U. Jo, B.~Lee, J.~Choi, H.~Yoo, and J.~Y. Choi, ``Cross-modal variational
  auto-encoder with distributed latent spaces and associators,'' \emph{arXiv
  preprint arXiv:1905.12867}, 2019.

\bibitem{7239600}
Y.~{Huang}, W.~{Wang}, and L.~{Wang}, ``Unconstrained multimodal multi-label
  learning,'' \emph{IEEE Transactions on Multimedia}, vol.~17, no.~11, pp.
  1923--1935, 2015.

\bibitem{zadeh-etal-2017-tensor}
\BIBentryALTinterwordspacing
A.~Zadeh, M.~Chen, S.~Poria, E.~Cambria, and L.-P. Morency, ``Tensor fusion
  network for multimodal sentiment analysis,'' in \emph{Proceedings of the 2017
  Conference on Empirical Methods in Natural Language Processing}.\hskip 1em
  plus 0.5em minus 0.4em\relax Copenhagen, Denmark: Association for
  Computational Linguistics, Sep. 2017, pp. 1103--1114. [Online]. Available:
  \url{https://www.aclweb.org/anthology/D17-1115}
\BIBentrySTDinterwordspacing

\bibitem{8752006}
S.~{Mai}, S.~{Xing}, and H.~{Hu}, ``Locally confined modality fusion network
  with a global perspective for multimodal human affective computing,''
  \emph{IEEE Transactions on Multimedia}, vol.~22, no.~1, pp. 122--137, 2020.

\bibitem{hadamard17}
\BIBentryALTinterwordspacing
J.~Kim, K.~W. On, W.~Lim, J.~Kim, J.~Ha, and B.~Zhang, ``Hadamard product for
  low-rank bilinear pooling,'' in \emph{5th International Conference on
  Learning Representations, {ICLR} 2017, Toulon, France, April 24-26, 2017,
  Conference Track Proceedings}, 2017. [Online]. Available:
  \url{https://openreview.net/forum?id=r1rhWnZkg}
\BIBentrySTDinterwordspacing

\bibitem{ben2017mutan}
H.~Ben-Younes, R.~Cadene, M.~Cord, and N.~Thome, ``Mutan: Multimodal tucker
  fusion for visual question answering,'' in \emph{Proceedings of the IEEE
  international conference on computer vision}, 2017, pp. 2612--2620.

\bibitem{zhang2020beyond}
X.~Zhang, X.~Gao, W.~Lu, L.~He, and J.~Li, ``Beyond vision: A multimodal
  recurrent attention convolutional neural network for unified image aesthetic
  prediction tasks,'' \emph{IEEE Transactions on Multimedia}, 2020.

\bibitem{long2018multimodal}
X.~Long, C.~Gan, G.~De~Melo, X.~Liu, Y.~Li, F.~Li, and S.~Wen, ``Multimodal
  keyless attention fusion for video classification,'' in \emph{Thirty-Second
  AAAI Conference on Artificial Intelligence}, 2018.

\bibitem{9206083}
F.~{Liu}, J.~{Liu}, Z.~{Fang}, R.~{Hong}, and H.~{Lu}, ``Visual question
  answering with dense inter- and intra-modality interactions,'' \emph{IEEE
  Transactions on Multimedia}, pp. 1--1, 2020.

\bibitem{tsai2019MULT}
Y.-H.~H. Tsai, S.~Bai, P.~P. Liang, J.~Z. Kolter, L.-P. Morency, and
  R.~Salakhutdinov, ``Multimodal transformer for unaligned multimodal language
  sequences,'' in \emph{Proceedings of the 57th Annual Meeting of the
  Association for Computational Linguistics (Volume 1: Long Papers)}.\hskip 1em
  plus 0.5em minus 0.4em\relax Florence, Italy: Association for Computational
  Linguistics, 7 2019.

\bibitem{degottex2014covarep}
G.~Degottex, J.~Kane, T.~Drugman, T.~Raitio, and S.~Scherer, ``Covarep—a
  collaborative voice analysis repository for speech technologies,'' in
  \emph{2014 IEEE international conference on acoustics, speech and signal
  processing (icassp)}.\hskip 1em plus 0.5em minus 0.4em\relax IEEE, 2014, pp.
  960--964.

\bibitem{drugman2011joint}
T.~Drugman and A.~Alwan, ``Joint robust voicing detection and pitch estimation
  based on residual harmonics,'' in \emph{Twelfth Annual Conference of the
  International Speech Communication Association}, 2011.

\bibitem{ghosh2016representation}
S.~Ghosh, E.~Laksana, L.-P. Morency, and S.~Scherer, ``Representation learning
  for speech emotion recognition.'' in \emph{Interspeech}, 2016, pp.
  3603--3607.

\bibitem{simonyan2014very}
K.~Simonyan and A.~Zisserman, ``Very deep convolutional networks for
  large-scale image recognition,'' in \emph{International Conference on
  Learning Representations}, 2015.

\bibitem{song-etal-2018-directional}
\BIBentryALTinterwordspacing
Y.~Song, S.~Shi, J.~Li, and H.~Zhang, ``Directional skip-gram: Explicitly
  distinguishing left and right context for word embeddings,'' in
  \emph{Proceedings of the 2018 Conference of the North {A}merican Chapter of
  the Association for Computational Linguistics: Human Language Technologies,
  Volume 2 (Short Papers)}.\hskip 1em plus 0.5em minus 0.4em\relax New Orleans,
  Louisiana: Association for Computational Linguistics, Jun. 2018, pp.
  175--180. [Online]. Available:
  \url{https://www.aclweb.org/anthology/N18-2028}
\BIBentrySTDinterwordspacing

\bibitem{pennington2014glove}
\BIBentryALTinterwordspacing
J.~Pennington, R.~Socher, and C.~D. Manning, ``Glove: Global vectors for word
  representation,'' in \emph{Empirical Methods in Natural Language Processing
  (EMNLP)}, 2014, pp. 1532--1543. [Online]. Available:
  \url{http://www.aclweb.org/anthology/D14-1162}
\BIBentrySTDinterwordspacing

\bibitem{wang2018non}
X.~Wang, R.~Girshick, A.~Gupta, and K.~He, ``Non-local neural networks,'' in
  \emph{Proceedings of the IEEE Conference on Computer Vision and Pattern
  Recognition}, 2018, pp. 7794--7803.

\bibitem{devlin2018bert}
\BIBentryALTinterwordspacing
J.~Devlin, M.-W. Chang, K.~Lee, and K.~Toutanova, ``{BERT}: Pre-training of
  deep bidirectional transformers for language understanding,'' in
  \emph{Proceedings of the 2019 Conference of the North {A}merican Chapter of
  the Association for Computational Linguistics: Human Language Technologies,
  Volume 1 (Long and Short Papers)}.\hskip 1em plus 0.5em minus 0.4em\relax
  Minneapolis, Minnesota: Association for Computational Linguistics, Jun. 2019,
  pp. 4171--4186. [Online]. Available:
  \url{https://www.aclweb.org/anthology/N19-1423}
\BIBentrySTDinterwordspacing

\bibitem{dumpala2019audio}
S.~H. Dumpala, R.~Chakraborty, and S.~K. Kopparapu, ``Audio-visual fusion for
  sentiment classification using cross-modal autoencoder,'' in \emph{32nd
  Conference on Neural Information Processing Systems (NIPS 2018)}, 2019, pp.
  1--4.

\bibitem{sun2016deep}
B.~Sun and K.~Saenko, ``Deep coral: Correlation alignment for deep domain
  adaptation,'' in \emph{European conference on computer vision}.\hskip 1em
  plus 0.5em minus 0.4em\relax Springer, 2016, pp. 443--450.

\bibitem{het-da}
\BIBentryALTinterwordspacing
J.~Li, K.~Lu, Z.~Huang, L.~Zhu, and H.~T. Shen, ``Heterogeneous domain
  adaptation through progressive alignment,'' \emph{{IEEE} Trans. Neural
  Networks Learn. Syst.}, vol.~30, no.~5, pp. 1381--1391, 2019. [Online].
  Available: \url{https://doi.org/10.1109/TNNLS.2018.2868854}
\BIBentrySTDinterwordspacing

\bibitem{mdd-da}
\BIBentryALTinterwordspacing
J.~Li, E.~Chen, Z.~Ding, L.~Zhu, K.~Lu, and H.~T. Shen, ``Maximum density
  divergence for domain adaptation,'' \emph{CoRR}, vol. abs/2004.12615, 2020.
  [Online]. Available: \url{https://arxiv.org/abs/2004.12615}
\BIBentrySTDinterwordspacing

\bibitem{localpreserve-da}
\BIBentryALTinterwordspacing
J.~Li, M.~Jing, K.~Lu, L.~Zhu, and H.~T. Shen, ``Locality preserving joint
  transfer for domain adaptation,'' \emph{{IEEE} Trans. Image Process.},
  vol.~28, no.~12, pp. 6103--6115, 2019. [Online]. Available:
  \url{https://doi.org/10.1109/TIP.2019.2924174}
\BIBentrySTDinterwordspacing

\bibitem{ijcai2019-626}
\BIBentryALTinterwordspacing
C.~Bai, S.~Kumar, J.~Leskovec, M.~Metzger, J.~F.~Nunamaker, and V.~S.
  Subrahmanian, ``Predicting the visual focus of attention in multi-person
  discussion videos,'' in \emph{Proceedings of the Twenty-Eighth International
  Joint Conference on Artificial Intelligence, {IJCAI-19}}.\hskip 1em plus
  0.5em minus 0.4em\relax International Joint Conferences on Artificial
  Intelligence Organization, 7 2019, pp. 4504--4510. [Online]. Available:
  \url{https://doi.org/10.24963/ijcai.2019/626}
\BIBentrySTDinterwordspacing

\bibitem{Marin19a}
M.~J. Marin-Jimenez, V.~Kalogeiton, P.~Medina-Suarez, and A.~Zisserman,
  ``{LAEO-Net}: revisiting people {Looking At Each Other} in videos,'' in
  \emph{International Conference on Computer Vision and Pattern Recognition
  (CVPR)}, 2019.

\bibitem{ioffe2015batch}
\BIBentryALTinterwordspacing
S.~Ioffe and C.~Szegedy, ``Batch normalization: Accelerating deep network
  training by reducing internal covariate shift,'' ser. Proceedings of Machine
  Learning Research, F.~Bach and D.~Blei, Eds., vol.~37.\hskip 1em plus 0.5em
  minus 0.4em\relax Lille, France: PMLR, 07--09 Jul 2015, pp. 448--456.
  [Online]. Available: \url{http://proceedings.mlr.press/v37/ioffe15.html}
\BIBentrySTDinterwordspacing

\bibitem{hinton2012improving}
G.~E. Hinton, N.~Srivastava, A.~Krizhevsky, I.~Sutskever, and R.~R.
  Salakhutdinov, ``Improving neural networks by preventing co-adaptation of
  feature detectors,'' \emph{arXiv preprint arXiv:1207.0580}, 2012.

\bibitem{kingma2014adam}
\BIBentryALTinterwordspacing
D.~P. Kingma and J.~Ba, ``Adam: {A} method for stochastic optimization,'' in
  \emph{3rd International Conference on Learning Representations, {ICLR} 2015,
  San Diego, CA, USA, May 7-9, 2015, Conference Track Proceedings}, Y.~Bengio
  and Y.~LeCun, Eds., 2015. [Online]. Available:
  \url{http://arxiv.org/abs/1412.6980}
\BIBentrySTDinterwordspacing

\bibitem{cer2018universal}
\BIBentryALTinterwordspacing
D.~Cer, Y.~Yang, S.-y. Kong, N.~Hua, N.~Limtiaco, R.~St.~John, N.~Constant,
  M.~Guajardo-Cespedes, S.~Yuan, C.~Tar, B.~Strope, and R.~Kurzweil,
  ``Universal sentence encoder for {E}nglish,'' in \emph{Proceedings of the
  2018 Conference on Empirical Methods in Natural Language Processing: System
  Demonstrations}.\hskip 1em plus 0.5em minus 0.4em\relax Brussels, Belgium:
  Association for Computational Linguistics, Nov. 2018, pp. 169--174. [Online].
  Available: \url{https://www.aclweb.org/anthology/D18-2029}
\BIBentrySTDinterwordspacing

\end{thebibliography}

\vspace{-10mm}
\begin{IEEEbiography}[{\includegraphics[width=1in,height=1in,clip,keepaspectratio]{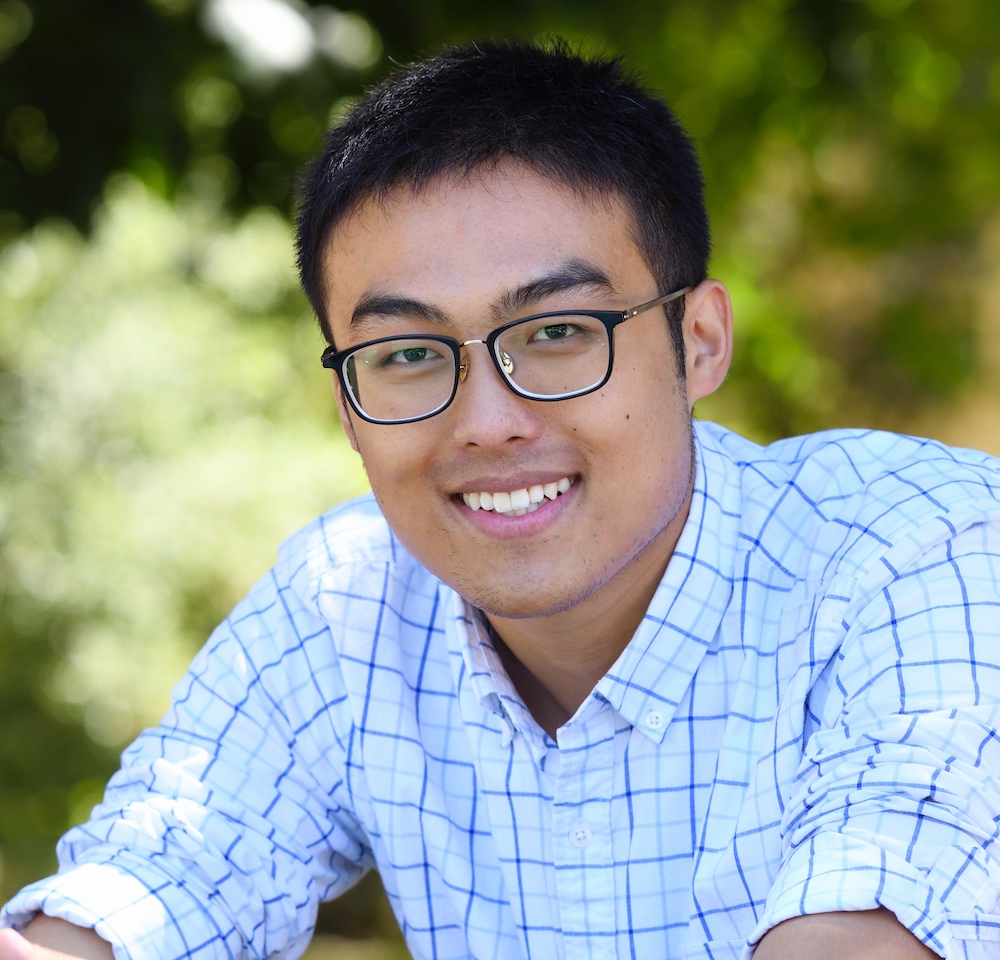}}]{Chongyang Bai}
is a Ph.D. candidate at Dartmouth College advised by Prof. V.S. Subrahmanian. He obtained a BS in Computational Mathematics and BEng in Computer Science from University of Science and Technology of China in 2016. He was a research intern in Microsoft Research and Google Research in 2016 and 2020 respectively. His research interests are multimodal learning and prediction as well as their applications to human behavioral analysis.
\end{IEEEbiography}
\vspace{-10mm}
\begin{IEEEbiography}[{\includegraphics[width=1in,height=1.25in,keepaspectratio]{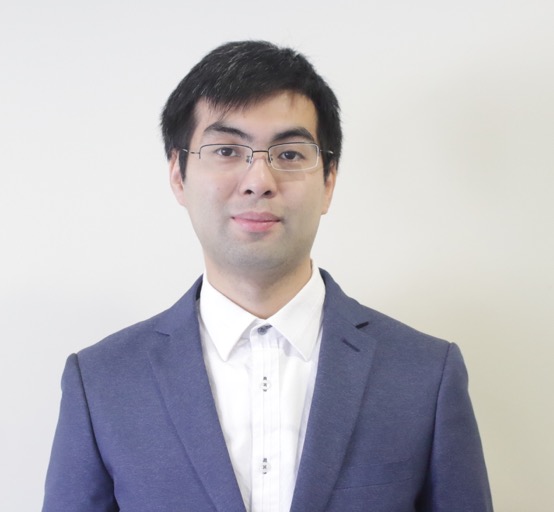}}]{Haipeng Chen}
is a postdoc in the Computer Science Department, Harvard University. Before that, he was a postdoc in the CS Department at Dartmouth College and obtained the PhD from Interdiscriplinary Graduate School, Nanyang Technological University, Singapore in 2018. His research lies in the general areas of Artificial Intelligence, including machine learning (reinforcement learning in particular), data mining, and algorithmic game theory, as well as their applications towards social good. He was winner for the 2017 Microsoft Malmo Collaborative AI Challenge, and runner-up for the Innovation Demonstration Award of IJCAI’19. He has published multiple papers in top tier conferences such as AAAI, IJCAI, UAI, KDD, ICDM. He serves as program committee member for top tier AI conferences such as Neurips, ICLR, AAAI, IJCAI and AAMAS.
\end{IEEEbiography}
\vspace{-10mm}
\begin{IEEEbiography}[{\includegraphics[width=1in,height=1.25in,clip,keepaspectratio]{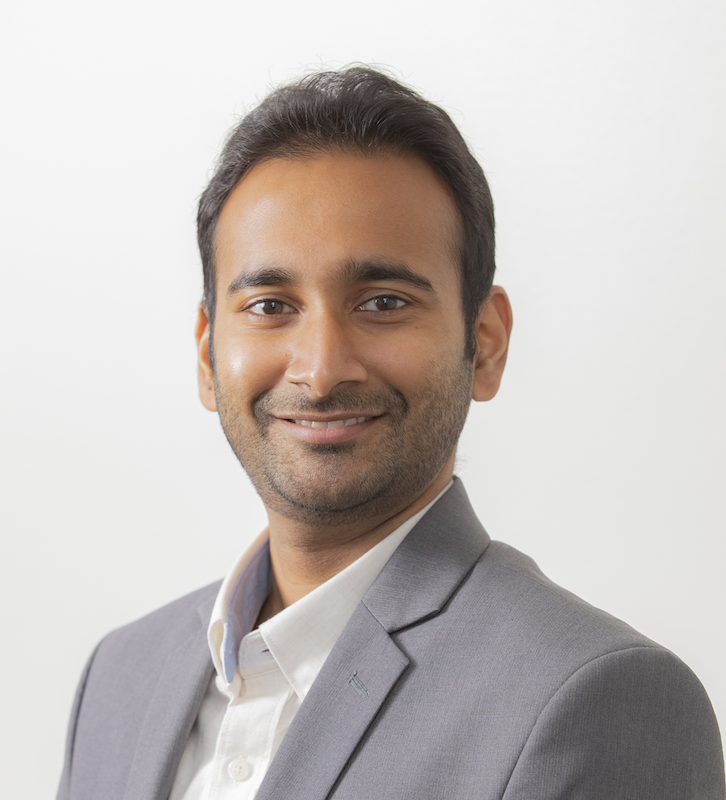}}]{Srijan Kumar} is an Assistant Professor at CSE, College of Computing at Georgia Institute of Technology. His research area is data science, machine learning, and network analytics. He has won several awards including Facebook Research Award, Adobe Faculty Research Award, ACM SIGKDD Doctoral Dissertation award runner-up, and best paper honorable mention at the World Wide Web Conference. He received his bachelor's degree in computer science from Indian Institute of Technology, masters and Ph.D. degree from the University of Maryland and completed his postdoctoral training from Stanford University. 
\end{IEEEbiography}
\vspace{-10mm}
\begin{IEEEbiography}[{\includegraphics[width=1in,height=1.25in,clip,keepaspectratio]{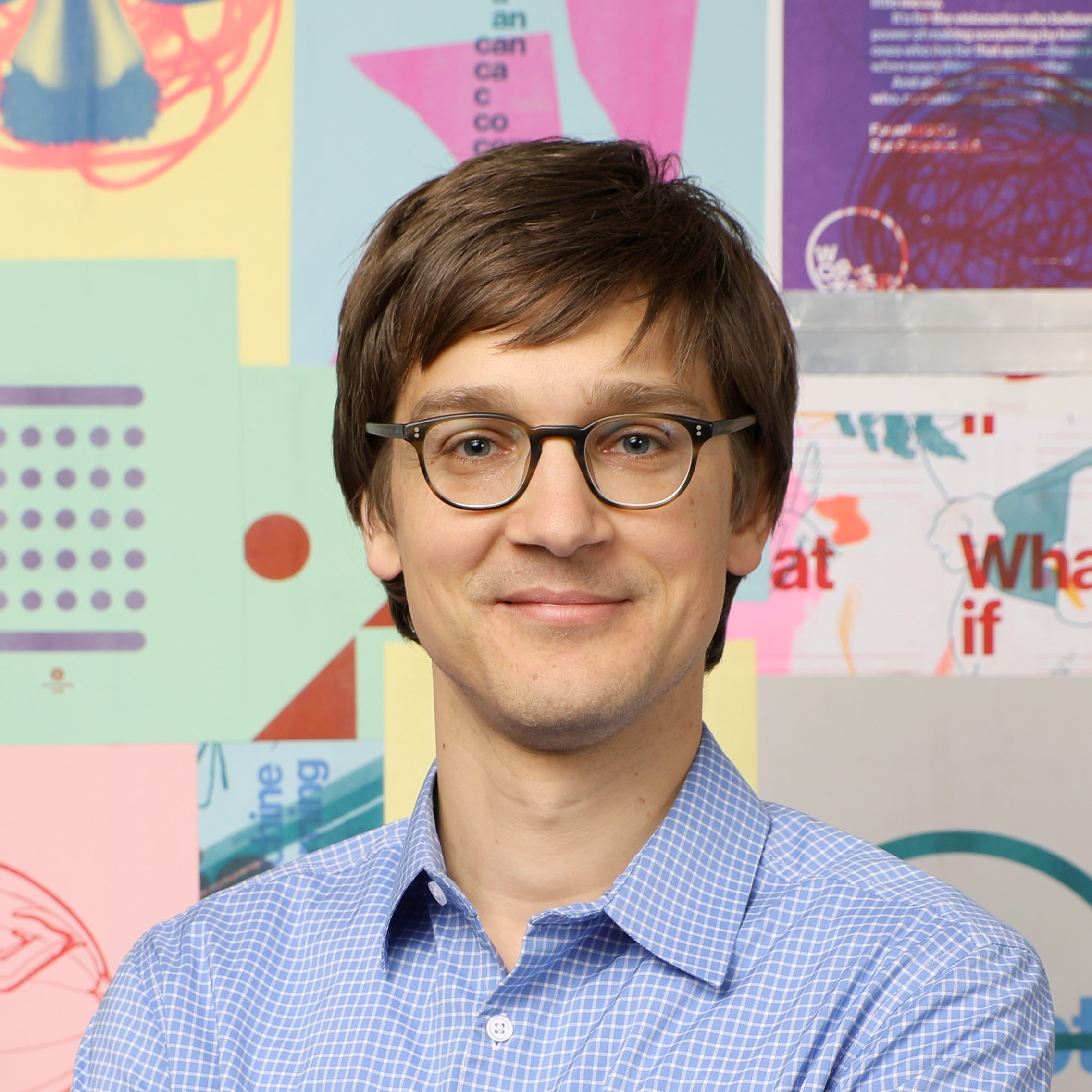}}]{Jure Leskovec}
is Associate Professor of Computer Science at Stanford University, Chief Scientist at Pinterest, and investigator at Chan Zuckerberg Biohub. His research focuses on machine learning and data mining with graphs, a general language for describing social, technological and biological systems. Computation over massive data is at the heart of his research and has applications in computer science, social sciences, marketing, and biomedicine. This research has won several awards including a Lagrange Prize, Microsoft Research Faculty Fellowship, the Alfred P. Sloan Fellowship, and numerous best paper and test of time awards. Leskovec received his bachelor’s degree in computer science from University of Ljubljana, Slovenia, PhD in machine learning from Carnegie Mellon University and postdoctoral training at Cornell University.
\end{IEEEbiography}
\vspace{-10mm}
\begin{IEEEbiography}[{\includegraphics[width=1in,height=1.25in,clip,keepaspectratio]{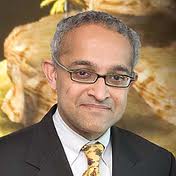}}]{V.S. Subrahmanian} is the Walter P. Murphy Professor of Computer Science and Buffett Faculty Fellow in the Buffett Institute for Global Affairs at Northwestern University. He previously served as the Dartmouth College Distinguished Professor in Cybersecurity, Technology, and Society and Director of the Institute for Security, Technology, and Society at Dartmouth and as a Professor of Computer Science at the University of Maryland from 1989-2017 where he also served for 6+ years as Director of the University of Maryland's Institute for Advanced Computer Studies. Prof. Subrahmanian is an expert on big data analytics including methods to analyze text/geospatial/relational/social network data, learn behavioral models from the data, forecast actions, and influence behaviors with applications to cybersecurity and counter-terrorism.  He has written eight books, edited fourteen, and published over 300 refereed articles. He is a Fellow of the American Association for the Advancement of Science and the Association for the Advancement of Artificial Intelligence and received numerous other honors and awards. His work has been featured in numerous outlets such as the Baltimore Sun, the Economist, Science, Nature, the Washington Post, American Public Media. He serves on the editorial boards of numerous journals including Science, the Board of Directors of SentiMetrix, Inc., and on the Research Advisory Board of Tata Consultancy Services. He previously served on Board of Directors of the Development Gateway, on DARPA's Executive Advisory Council on Advanced Logistics and as an ad-hoc member of the US Air Force Science Advisory Board.
\end{IEEEbiography}







\end{document}